\pdfoutput=1

\documentclass[11pt]{article}

\usepackage{coling}

\usepackage{times}
\usepackage{latexsym}

\usepackage[T1]{fontenc}

\usepackage[utf8]{inputenc}

\usepackage{microtype}

\usepackage{inconsolata}

\usepackage{graphicx}
\usepackage{algorithm}
\usepackage{algorithmic}
\usepackage{amsmath}
\usepackage{amssymb}
\usepackage{multirow}
\usepackage{makecell}
\usepackage{xcolor}

\title{MiMoTable: A Multi-scale Spreadsheet Benchmark with \\Meta Operations for Table Reasoning}

\author{Zheng Li$^*$, Yang Du$^*$, Mao Zheng$^*$, Mingyang Song\thanks{Equal contribution.}\\
	Tencent Hunyuan \\
	{\tt jasonzli@tencent.com} \\
}

\begin{document}
\maketitle
\begin{abstract}
Extensive research has been conducted to explore the capability of Large Language Models (LLMs) for table reasoning and has significantly improved the performance on existing benchmarks. However, tables and user questions in real-world applications are more complex and diverse, presenting an unignorable gap compared to the existing benchmarks. To fill the gap, we propose a \textbf{M}ult\textbf{i}-scale spreadsheet benchmark with \textbf{M}eta \textbf{o}perations for \textbf{Table} reasoning, named as MiMoTable. Specifically, MiMoTable incorporates two key features. First, the tables in MiMoTable are all spreadsheets used in real-world scenarios, which cover seven domains and contain different types. Second, we define a new criterion with six categories of meta operations for measuring the difficulty of each question in MiMoTable, simultaneously as a new perspective for measuring the difficulty of the existing benchmarks. Experimental results show that Claude-3.5-Sonnet achieves the best performance with 77.4\% accuracy, indicating that there is still significant room to improve for LLMs on MiMoTable. Furthermore, we grade the difficulty of existing benchmarks according to our new criteria. Experiments have shown that the performance of LLMs decreases as the difficulty of benchmarks increases, thereby proving the effectiveness of our proposed new criterion. All data and code are open-sourced at \url{https://github.com/jasonNLP/MiMoTable}.
\end{abstract}

%

\section{Introduction}
Tabular data plays a crucial role across diverse domains, including education, finance, and others. Table reasoning involves deriving meaningful insights and answers from structured tabular data to address specific user queries \cite{survey_table_reasoning}. This process significantly improves the efficiency of information retrieval and interpretation for users. To foster a comprehensive understanding of this field, researchers have proposed and developed numerous table reasoning tasks, such as TableQA, Table2Text, Table Manipulation, and Advanced Data Analysis \cite{Survey}. Various methods have been proposed to tackle these tasks, and large language models (LLMs) have achieved promising results \cite{TAPEX, Binder}. To evaluate performance, several table reasoning benchmarks have been introduced, including WikiTableQuestions \cite{WikiTableQuestion}, ToTTo \cite{ToTTo}, SheetCopilot \cite{SheetCopilot}, Text2Analysis \cite{Text2Analysis} and so on.

\begin{figure}[t!]
\centering
\includegraphics[scale=0.64]{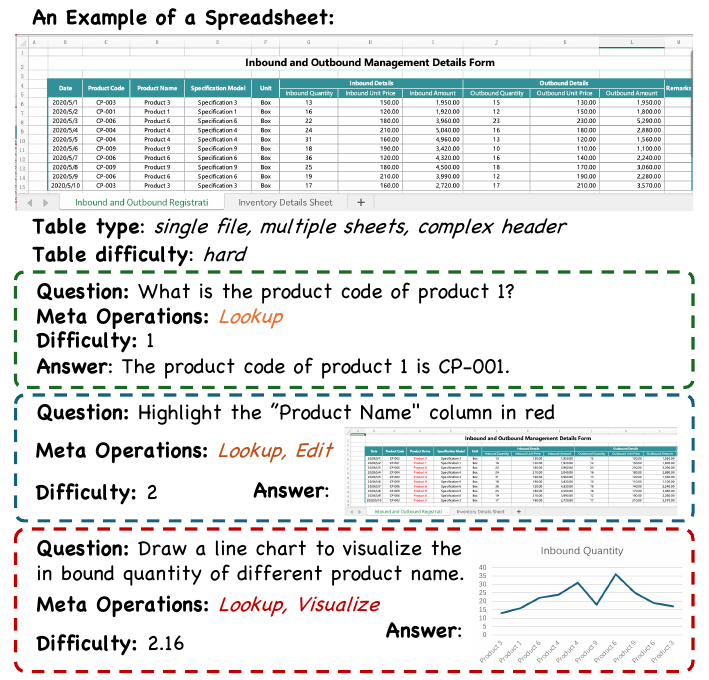}
\caption{Examples of MiMoTable benchmark.}
\label{example_of_mimo}
\end{figure}

In the realm of table reasoning, benchmark development has not kept pace with the rapid advancements in methodological approaches. While LLMs have exhibited remarkable performance on existing benchmarks, recent studies have brought to light persistent limitations in their capacity for nuanced table comprehension \cite{SUC}. Upon critical analysis, we have identified shortcomings in existing benchmarks across two key aspects.

First, current benchmarks exhibit significant limitations in their representation of real-world tabular data complexity. Most tables in these benchmarks have simple headers (single-row/column) and fail to cover all four task types comprehensively.   However, real-world tables are diverse and can be divided into three parts: 1) Headers ranging from single-row/column to complex hierarchical forms. 2) Variable number of sheets in Excel files. 3) Multiple tables within a single sheet. These complexities are often overlooked in existing benchmarks, limiting their ability to accurately assess table reasoning capabilities in practical applications. 

\begin{figure*}[t]
\centering
\includegraphics[scale=0.234]{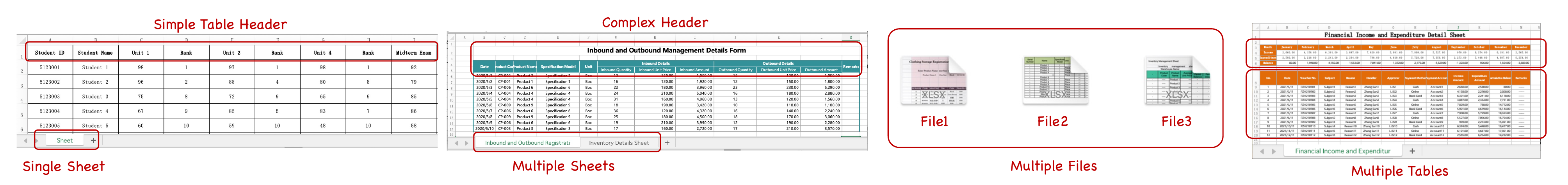}
\caption{Illustrations of different table types, including simple header, complex header, single sheet, multiple sheets, multiple files, and multiple tables in one sheet.}
\label{table_types}
\end{figure*}

Second, although current existing benchmarks are divided according to task granularity, the difficulty of different benchmark datasets within the same task can vary. For example, the WikiSQL \cite{WikiSQL} dataset is simpler than the unrestricted WikiTableQuestions dataset because it limits questions to those that can be answered using a subset of SQL queries. The current task divisions cannot reflect this difference in difficulty.

To address the above issues, we propose MiMoTable, a table reasoning benchmark with diverse spreadsheets and meta operations. Our dataset comprises 428 spreadsheets from real-world scenarios, spanning seven domains: architecture, finance, office, education, accounting, e-commerce, and manufacturing. Our table data is comprehensive, featuring both simple and complex headers, and varying in the number of sheets from single to multiple. Some spreadsheets even contain multiple tables within a single sheet. We have constructed 1,719 question-answer pairs based on these spreadsheets, forming (spreadsheet, question, answer) triplets. Examples of MiMoTable are shown in Figure \ref{example_of_mimo}.

Simultaneously, to more deeply reflect the differences in dataset problems, we propose a new criterion for categorizing problems based on meta operations. There are six types of meta operations: Lookup, Edit, Compare, Calculate, Visualize, and Reasoning. Each type of meta operation corresponds to a difficulty score. With the new criterion, we can associate each problem with one or more meta operations, thereby assigning a difficulty score to each problem. In this way, different benchmarks can be graded into different difficulty scores, facilitating better analysis and comparison. 

\begin{table}[t]
\centering
\scriptsize
\renewcommand\tabcolsep{11pt}
\renewcommand\arraystretch{1.7}
\begin{tabular}{c|c|c} \hline
     & \textbf{Simple Header} & \textbf{Complex Header} \\ \hline
    \textbf{Single Sheet} & \textit{simple table} & {\color{blue}\textit{medium table}} \\ \hline
    \textbf{Multiple Sheets} & {\color{blue}\textit{medium table}} & {\color{red}\textit{hard table}} \\ \hline
    \textbf{Multiple Files} & {\color{blue}\textit{medium table}} & {\color{red}\textit{hard table}} \\ \hline
    \textbf{Multiple Tables} & {\color{red}\textit{hard table}} & {\color{red}\textit{hard table}} \\ \hline
\end{tabular}
\caption{Categories of table difficulty.}
\label{cat_of_table_difficulty}
\end{table}

\begin{table*}[t]
\scriptsize
\centering
\renewcommand\tabcolsep{13pt}
\renewcommand\arraystretch{1.7}
\begin{tabular}{c|c|c|c} \hline
    \textbf{Meta Operations} & \textbf{Description} & \textbf{Grade} & \textbf{Examples}\\ \hline
    Lookup & Locate the position of specific target & 1 & What is the product code of product 1? \\ \hline
    Edit & Modify, delete or add in a table & 1 & Highlight the ``Product Name" column in red \\ \hline
    Calculate & The numerical computation, sum, avg, max, etc & 2 & How many students are in the table? \\ \hline
    Compare & Compare two or more targets in a table & 2 & Who has the highest score? \\ \hline
    Visualize & Show in chart & 2 & Draw a chart to show the distribution of scores. \\ \hline
    Reasoning & \makecell{Inferring information from the table \\content that is not explicitly included} & 3 & \makecell{Analyze the relationship between \\the loan term, monthly interest, and interest.} \\ \hline
\end{tabular}
\caption{The description and grade of meta operations.}
\label{mean_of_meta_op}
\end{table*}
\begin{figure}[t!]
\centering
\includegraphics[width=1\columnwidth]{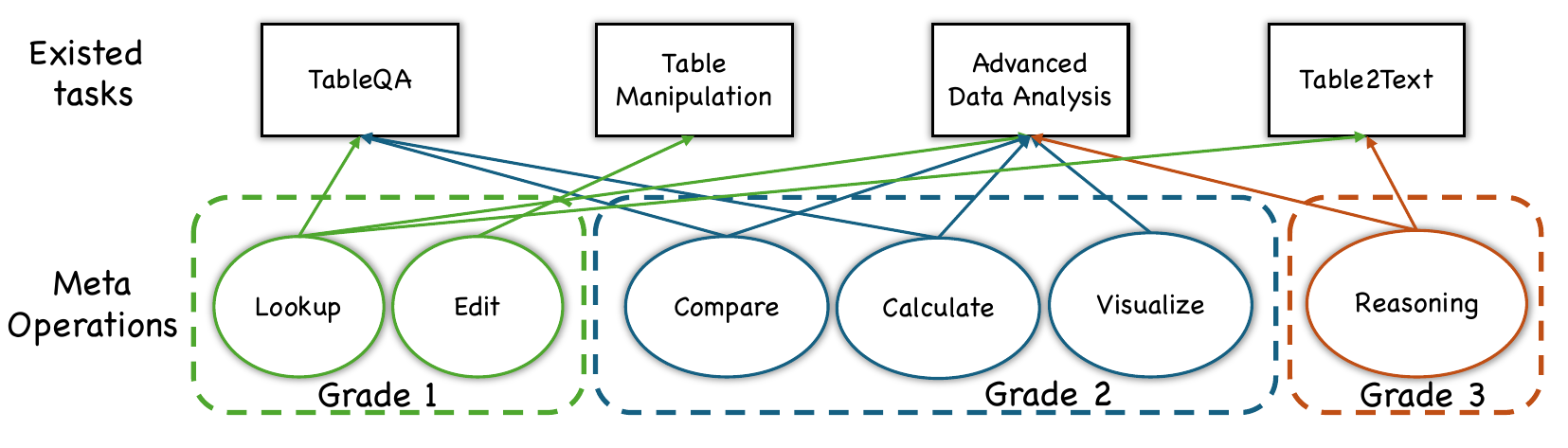}
\caption{The relationships between tasks in the existing benchmarks and our proposed meta operations.}
\label{mapping}
\end{figure}

Our main contributions are as follows:
\begin{itemize}
\item We propose a new benchmark comprising 428 multi-scale spreadsheets in both Chinese and English, featuring simple and complex headers, single and multiple sheets, single and multiple files, and multiple tables within a single sheet. Based on these characteristics, we classify the tables into three difficulty levels: simple, medium, and hard. We construct 1,719 (spreadsheet, question, answer) triplets covering a wide range of tasks. 

\item We introduce a novel criterion for categorizing table reasoning problems using meta operations, each assigned a difficulty score. These non-overlapping meta operations can be combined to represent existing tasks, allowing for a more precise evaluation of a model's capabilities in table-related tasks.

\item We conducted extensive experiments, demonstrating that the proposed benchmark is challenging for existing LLMs and proving the effectiveness of the proposed meta operations.
\end{itemize}
\section{MiMoTable Benchmark}

In this section, we introduce how to prepare our new MiMoTable benchmark and ensure its quality.

\subsection{Types and Difficulty of Tables}

Most existing table reasoning benchmarks utilize single tables with simple headers, contrasting with the diverse tables encountered in real-world scenarios, particularly in spreadsheets like Excel files. After analyzing real-world spreadsheets, we categorize them along four dimensions: header types, the number of sheets per file, the number of tables per sheet, and the file count.

As shown in Figure \ref{table_types}, header types can be divided into simple headers and complex headers. A simple header refers to a single-row or single-column header, while all others are considered complex headers. For example, hierarchical headers are classified as complex headers. Understanding complex headers is more challenging than understanding simple ones, and multiple sheets generally contain more information than a single sheet. Therefore, based on the aforementioned dimensions, we classify spreadsheets into three difficulty levels: simple, medium, and hard. The specific classification rules are illustrated in Table \ref{cat_of_table_difficulty}.

\subsection{Meta Operations}

Current table reasoning benchmarks are classified by tasks, mainly including TableQA, Table2Text, Table Manipulation, and Advanced Data Analysis. This task-based categorization evaluates model performance across different tasks but fails to measure differences between benchmarks within the same task or compare benchmarks from different tasks along the same dimension.

To enhance the analysis of table reasoning benchmarks, we propose a novel criterion categorizing questions by meta operations: Lookup, Edit, Calculate, Compare, Visualize, and Reasoning. Table \ref{mean_of_meta_op} defines each operation. These operations reflect specific LLM capabilities in handling table-related problems, with questions potentially involving multiple operations. For instance, "How many students are in the table" requires both Lookup (locating student names) and Calculate (counting them) operations.

The combination of six meta operations can encompass tasks in current table benchmarks. Figure \ref{mapping} shows the mapping relationship between existing tasks and meta operations. For instance, TableQA questions may involve combinations of Lookup, Compare, and Calculate operations.

Additionally, to assess problem complexity, we categorize the six meta operations into three difficulty grades (1, 2, 3) based on common criteria, as shown in Table \ref{mean_of_meta_op}. Lookup and Edit, involving simple content location or modification, are grade 1. Compare, Calculate, and Visualize, which require logical operations, are grade 2. Reasoning, necessitating inference beyond explicit table content, is the most complex at grade 3.

With the difficulty score of meta operations, we can calculate the difficulty score of each question and the entire dataset. First, let's assume there are N questions in the dataset. The $i$-th problem $q_{i}$ can be associated with $K_{i}$ meta operations. Suppose the $k$-th meta operation is denoted as $op_{k}$, then the sequence of meta operations for the $i$-th questions can be represented as:
\begin{figure*}[ht]
\centering
\includegraphics[width=1\linewidth]{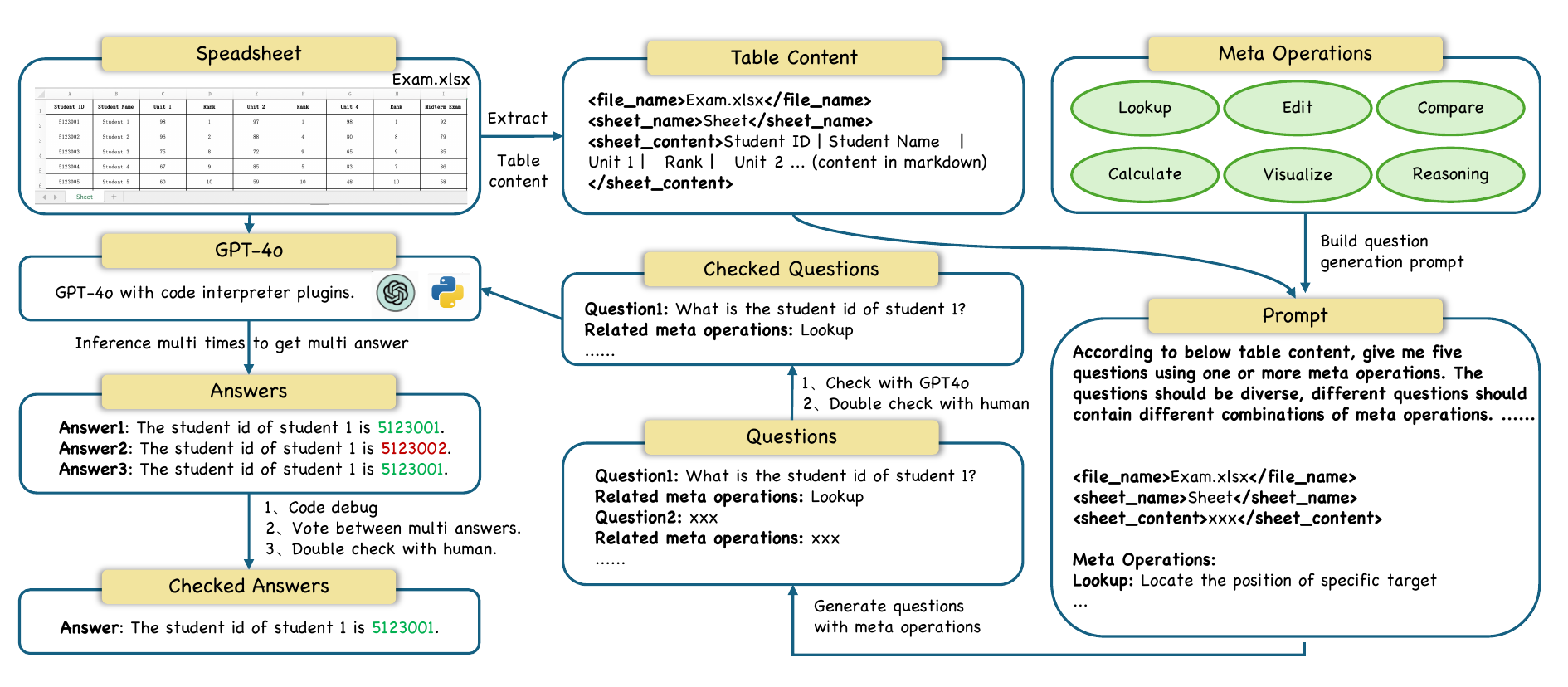}
\caption{The data construction pipeline of MiMoTable benchmark.}
\label{pipline}
\end{figure*}
\begin{equation}
OP_{q_i}=[op_1, op_2, ... , op_{K_i}]
\end{equation}
The difficulty sequence corresponding to the meta operations can be represented as:
\begin{equation}
S_{q_i}=[s_1, s_2, ... , s_{K_i}], s_k \in{1,2, 3}
\end{equation}
where $s_{K_i}$ indicates the difficulty score corresponding to meta operations $op_{k}$.

To ensure that questions involving more difficult meta operations are assigned a higher difficulty score, we define the difficulty score of a question $q_{i}$ as follows:
\begin{equation}
qs_i=ms_{q_i}+(\sum\limits_1^{K_i} s_i-ms_{q_i})/M_{ms_{q_i}}
\label{cal_of_q_d}
\end{equation}
\begin{equation}
ms_{q_i}=\max(S_{q_i})
\end{equation}

The term $M_{ms_{q_i}}$ refers to the maximum value that $\sum\limits_1^{K_i} s_i-ms_{q_i}$ can be achieved under the condition where the meta operations of the question have the highest level of difficulty $ms_{q_i}$. Every meta operation can only appear once in each question. So when the $ms_{q_i} = 3$, which means the corresponding meta operation is Reasoning, the most complex combination of the rest of the meta operations is Compare, Calculate, Visualize, Lookup, and Edit. The difficulty score sum of those meta operations is 2 + 2 + 2 + 1 + 1 = 8. So $M_3$=8. And in the same manner, we can get all the values of $M_{ms_{q_i}}$ as follows, 
\begin{equation}
M_{ms_{q_i}}= \begin{cases}1 & ms_{q_i} = 1\\6 & ms_{q_i} = 2\\8 & ms_{q_i} = 3\end{cases}
\end{equation}
So the range of $qs_i$ is [1, 4]. With the difficulty score $qs_i$ of a single question $q_{i}$, we define the difficulty score of the entire dataset, $ds$, as the average of the difficulty scores of all questions:
\begin{equation}
ds=\frac{\sum_{1}^{N}qs_i}{N}
\label{score of dataset}
\end{equation}

\begin{table*}[h]
\centering
\tiny
\renewcommand\tabcolsep{3pt}
\renewcommand\arraystretch{1.5}
\begin{tabular}{c|cccc|cccc}
\hline
\multirow{2}{*}{\textbf{Benchmarks}} & \multicolumn{4}{c|}{\textbf{Table   Types}} & \multicolumn{4}{c}{\textbf{Tasks}} \\ \cline{2-9} 
 & \multicolumn{1}{c|}{\textbf{Header Type}} & \multicolumn{1}{c|}{\textbf{Sheet Num}} & \multicolumn{1}{c|}{\textbf{File Num}} & \textbf{Table Num} & \multicolumn{1}{c|}{\textbf{TableQA}} & \multicolumn{1}{c|}{\textbf{Table2Text}} & \multicolumn{1}{c|}{\textbf{Table Manipulation}} & \textbf{Advanced Data   Analysis} \\ \hline
WikiTableQuestion & \multicolumn{1}{c|}{simple} & \multicolumn{1}{c|}{single} & \multicolumn{1}{c|}{single} & single & \multicolumn{1}{c|}{\checkmark} & \multicolumn{1}{c|}{} & \multicolumn{1}{c|}{} &  \\ \hline
WikiSQL & \multicolumn{1}{c|}{simple} & \multicolumn{1}{c|}{single} & \multicolumn{1}{c|}{single} & single & \multicolumn{1}{c|}{\checkmark} & \multicolumn{1}{c|}{} & \multicolumn{1}{c|}{} &  \\ \hline
FetaQA & \multicolumn{1}{c|}{simple} & \multicolumn{1}{c|}{single} & \multicolumn{1}{c|}{single} & single & \multicolumn{1}{c|}{\checkmark} & \multicolumn{1}{c|}{} & \multicolumn{1}{c|}{} &  \\ \hline
HiTAB & \multicolumn{1}{c|}{complex} & \multicolumn{1}{c|}{single} & \multicolumn{1}{c|}{single} & single & \multicolumn{1}{c|}{\checkmark} & \multicolumn{1}{c|}{\checkmark} & \multicolumn{1}{c|}{} &  \\ \hline
ToTTo & \multicolumn{1}{c|}{simple} & \multicolumn{1}{c|}{single} & \multicolumn{1}{c|}{single} & single & \multicolumn{1}{c|}{} & \multicolumn{1}{c|}{\checkmark} & \multicolumn{1}{c|}{} &  \\ \hline
DAEval & \multicolumn{1}{c|}{simple} & \multicolumn{1}{c|}{single} & \multicolumn{1}{c|}{single} & single & \multicolumn{1}{c|}{} & \multicolumn{1}{c|}{} & \multicolumn{1}{c|}{} & \checkmark \\ \hline
WikiTableEdit & \multicolumn{1}{c|}{simple} & \multicolumn{1}{c|}{single} & \multicolumn{1}{c|}{single} & single & \multicolumn{1}{c|}{} & \multicolumn{1}{c|}{} & \multicolumn{1}{c|}{\checkmark} &  \\ \hline
Text2Analysis & \multicolumn{1}{c|}{simple} & \multicolumn{1}{c|}{single} & \multicolumn{1}{c|}{single} & single & \multicolumn{1}{c|}{\checkmark} & \multicolumn{1}{c|}{\checkmark} & \multicolumn{1}{c|}{} & \checkmark \\ \hline
MiMoTable(ours) & \multicolumn{1}{c|}{simple \& complex} & \multicolumn{1}{c|}{single \& multiple} & \multicolumn{1}{c|}{single \& multiple} & single \& multiple & \multicolumn{1}{c|}{\checkmark} & \multicolumn{1}{c|}{\checkmark} & \multicolumn{1}{c|}{\checkmark} & \checkmark \\ \hline
\end{tabular}
\caption{Comparison in table types and tasks between existing benchmarks and MiMoTable}
\label{compare_in_tabel_types}
\end{table*}

\subsection{Dataset Construction}
We introduce how the dataset is constructed from three aspects: table collection, question generation, and answer generation. Figure \ref{pipline} illustrates the whole construction process.

\noindent\textbf{Table Collection.} Since Excel files are the most popular spreadsheet in real-world scenarios, we choose .xlsx as the file format to be collected. The spreadsheets of our dataset are collected from publicly available sources on the internet. The Chinese tables primarily come from Baidu Wenku, while the English tables are mainly sourced from Google searches. These spreadsheets cover seven common domains: architecture, finance, office, education, accounting, e-commerce, and manufacturing. To ensure that the types of spreadsheets encompass as many real-world scenarios as possible, according to the classifications of table type mentioned before, we collected spreadsheets with both simple and complex headers, as well as those with single and multiple sheets. Even within a single sheet, our data may contain multiple tables. Additionally, we randomly sampled some individual spreadsheet files and combined them into groups of 2-5 files, and the subsequent questions and answers are generated with those multiple files as input. To maintain the quality of the collected spreadsheets, we manually checked the content, removing files with significant noise, garbled text, or non-tabular formats. Furthermore, we reviewed each spreadsheet to anonymize any potential private information. Specifically: (1) personal names, contact information, addresses, etc., are masked and randomly regenerated by GPT, while headers are retained due to their importance and generality; (2) we double-checked the final spreadsheets with legal professionals.

Ultimately, we obtained 428 high-quality spreadsheets containing both Chinese and English languages and various types. As shown in Table \ref{compare_in_tabel_types}, compared to the current benchmarks, our collected spreadsheets far exceed in diversity of types, better reflecting various real-world scenarios.

\noindent\textbf{Question Generation.} As shown in Figure \ref{pipline}, we use GPT-4o to generate relevant questions and double-check with models and humans. First, we extract the table content from a spreadsheet in markdown format. Then, according to the extracted table content and our meta operations, GPT-4o is prompted to generate related questions. We instructed the model to generate multiple questions at once for each spreadsheet. To ensure the diversity of questions, the multiple questions should contain different combinations of meta operations. For multi-sheet or multi-file spreadsheets, we prompt the model to generate questions requiring cross-sheet or cross-file analysis. To ensure prompt effectiveness, we initially generate 50 samples, conduct a human evaluation to identify issues, and iteratively refine the prompt until most generated questions meet our criteria.

After generating initial questions with GPT-4o, we prompt it to verify if they meet requirements: relevance to table content and correct meta operations. We then manually review and filter out unsuitable questions, yielding 1,719 high-quality, comprehensive questions.

We classify the questions by existing tasks, covering TableQA, Table2Text, Table Manipulation, and Advanced Data Analysis. As Table \ref{compare_in_tabel_types} shows, our dataset exceeds all current table benchmarks in task comprehensiveness.

\begin{figure}[h]
\centering
\includegraphics[width=0.7\columnwidth]{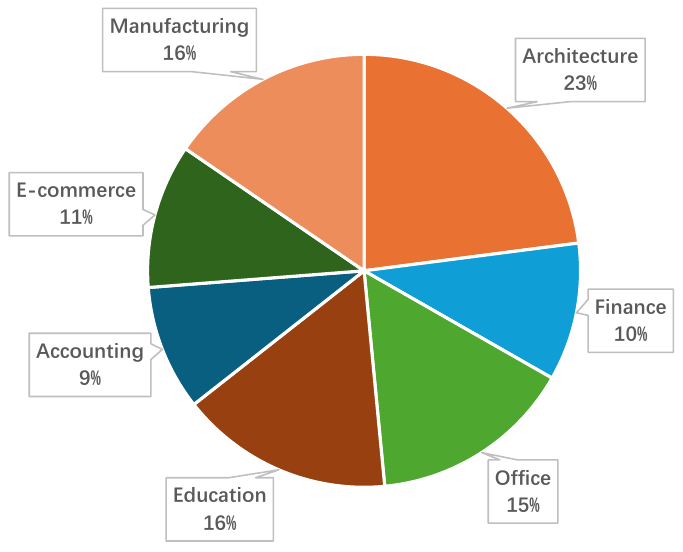}
\caption{Domain distribution of all spreadsheets.}
\label{Domain Distribution}
\end{figure}

\noindent\textbf{Answer Generation.} After we collected the tables and generated the related questions based on meta operations, the final step is to obtain the corresponding answers. Since some questions are related to editing on the origin spreadsheet or drawing charts, we leverage GPT-4o with the code interpreter plugin to get initial answers. The spreadsheet files can be directly used as inputs, and the model can generate Python code to run in a code interpreter to generate the modified files or visual charts.

As shown in Figure \ref{pipline}, we ensure GPT-4o answer quality by first debugging its code. If the code cannot be executed without errors, the answer is considered incorrect. Second, We perform multiple inferences on each question-table pair, selecting the most frequent answer as a candidate. If all answers are different, the sample is viewed as invalid due to the inconsistency. Last, we have table analysis experts manually annotate the candidate answers, retaining the correct ones and correcting the wrong ones. We invited 10 experts with data analysis experience to annotate the dataset. Among them, native Chinese and English speakers each accounted for half of the group. Each answer is annotated twice, and Cohen's Kappa is 0.83, which indicates a high inter-annotator agreement. The answers of our dataset not only contain text but also contain Excel files and charts.

\subsection{Dataset Statistic}
Our MiMoTable benchmark consists of 1,719 (spreadsheet, question, answer) triplets originating from 428 different spreadsheets. In this subsection, we provide statistics from different dimensions to provide a more comprehensive understanding of our dataset.

\noindent\textbf{Domains of Spreadsheets.} As illustrated in Figure \ref{Domain Distribution}, our spreadsheets encompass seven domains in real-world applications.

\noindent\textbf{Type and Difficulty of Tables.} From Table \ref{stat_of_table_type}, we can see that the table type of the collected spreadsheet is diverse, covering both simple, medium, and hard difficulty. 

\begin{table}[h]
\centering
\scriptsize
\renewcommand\tabcolsep{1.5pt}
\renewcommand\arraystretch{1.5}
\begin{tabular}{c|c|c|c}
\hline
\textbf{Difficulty} & \textbf{Ratio} & \textbf{Table Type} & \textbf{Num} \\ \hline
Simple & 33.6\% & single file + single sheet + simple header & 144 \\ \hline
\multirow{3}{*}{Medium} & \multirow{3}{*}{32.5\%} & single file + multiple sheets + simple header & 30 \\ \cline{3-4} 
 &  & multiple files + simple header & 37 \\ \cline{3-4} 
 &  & single file + single sheet + complicate header & 72 \\ \hline
\multirow{3}{*}{Hard} & \multirow{3}{*}{33.9\%} & single file + multiple sheets + complicate header & 63 \\ \cline{3-4} 
 &  & multiple files + complicate header & 32 \\ \cline{3-4} 
 &  & multiple tables & 50 \\ \hline
\end{tabular}
\caption{Distribution of table difficulty.}
\label{stat_of_table_type}
\end{table}

\begin{figure}[h]
\centering
\includegraphics[width=1\columnwidth]{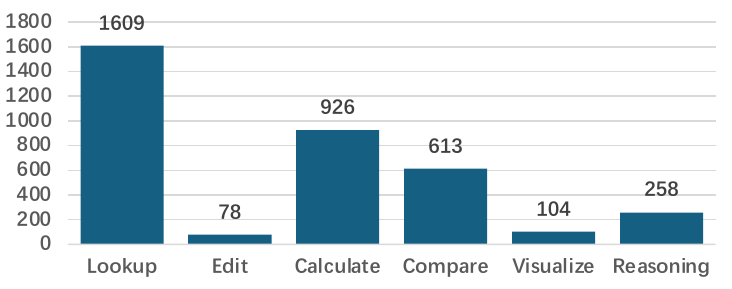}
\caption{Distribution of meta operations.}
\label{meta_op_stat}
\end{figure}

\noindent\textbf{Meta Operations of Questions.} Figure \ref{meta_op_stat} shows the number of six meta operations in our benchmark questions. As the most basic operation of a table, Lookup is the most frequently occurring meta operation. More difficult meta operations such as Calculate and Reasoning also account for a relatively large proportion of our dataset, indicating that the questions of MiMoTable are diverse and comprehensive.

\noindent\textbf{Difficulty of Questions.} To investigate the difficulty of questions, we calculate the difficulty score of each question according to the Equation
 \ref{cal_of_q_d}. The score is in the range of [1, 4], so we divided the distribution of scores into three intervals, [1, 2), [2, 3) and [3, 4]. The specific values of question number and ratio are in Table \ref{stat_of_question_difficulty}.

\begin{table}[h]
\centering
\scriptsize
\renewcommand\tabcolsep{20pt}
\renewcommand\arraystretch{1.2}
\begin{tabular}{ccc}
\hline
\textbf{Question Difficulty} & \textbf{Num} & \textbf{Ratio} \\ \hline
{[}1, 2) & 311 & 18.1\% \\
{[}2, 3) & 1150 & 66.9\% \\
{[}3, 4] & 258 & 15.0\% \\ \hline
\end{tabular}
\caption{Distribution of question difficulty.}
\label{stat_of_question_difficulty}
\end{table}

\begin{table*}[t!]
\centering
\scriptsize
\renewcommand\tabcolsep{1.7pt}
\renewcommand\arraystretch{1.6}
\begin{tabular}{c|c|cc|ccc|ccc|cccc}
\hline
\multirow{2}{*}{\textbf{Model}} & \multirow{2}{*}{\textbf{Overall}} & \multicolumn{2}{c|}{\textbf{Language}} & \multicolumn{3}{c|}{\textbf{Table Difficulty}} & \multicolumn{3}{c|}{\textbf{Question Difficulty}} & \multicolumn{4}{c}{\textbf{Meta Operations}} \\ \cline{3-14} 
 &  & \multicolumn{1}{c|}{\textbf{English}} & \textbf{Chinese} & \multicolumn{1}{c|}{\textbf{Simple}} & \multicolumn{1}{c|}{\textbf{Medium}} & \textbf{Hard} & \multicolumn{1}{c|}{\textbf{{[}1, 2)}} & \multicolumn{1}{c|}{\textbf{{[}2, 3)}} & \textbf{{[}3, 4{]}} & \multicolumn{1}{c|}{\textbf{Lookup}} & \multicolumn{1}{c|}{\textbf{Compare}} & \multicolumn{1}{c|}{\textbf{Calculate}} & \textbf{Reasoning} \\ \hline
Claude-3.5-Sonnet & \textbf{77.4\%} & \multicolumn{1}{c|}{\textbf{79.0\%}} & \textbf{76.2\%} & \multicolumn{1}{c|}{81.3\%} & \multicolumn{1}{c|}{\textbf{75.5\%}} & \textbf{72.1\%} & \multicolumn{1}{c|}{\textbf{89.0\%}} & \multicolumn{1}{c|}{\textbf{77.1\%}} & \textbf{63.3\%} & \multicolumn{1}{c|}{\textbf{89.0\%}} & \multicolumn{1}{c|}{\textbf{79.7\%}} & \multicolumn{1}{c|}{\textbf{76.1\%}} & \textbf{63.3\%} \\ \hline
GPT-4o-CI & 69.2\% & \multicolumn{1}{c|}{70.8\%} & 68.1\% & \multicolumn{1}{c|}{\textbf{81.7\%}} & \multicolumn{1}{c|}{67.1\%} & 50.8\% & \multicolumn{1}{c|}{81.0\%} & \multicolumn{1}{c|}{71.1\%} & 45.8\% & \multicolumn{1}{c|}{81.0\%} & \multicolumn{1}{c|}{73.1\%} & \multicolumn{1}{c|}{70.6\%} & 45.8\% \\ \hline
GPT-4o-TXT & 69.0\% & \multicolumn{1}{c|}{69.3\%} & 68.8\% & \multicolumn{1}{c|}{73.8\%} & \multicolumn{1}{c|}{66.2\%} & 62.1\% & \multicolumn{1}{c|}{85.1\%} & \multicolumn{1}{c|}{67.6\%} & 53.9\% & \multicolumn{1}{c|}{85.1\%} & \multicolumn{1}{c|}{73.1\%} & \multicolumn{1}{c|}{64.5\%} & 53.9\% \\ \hline
Gemini-1.5-Pro & 60.2\% & \multicolumn{1}{c|}{61.6\%} & 59.1\% & \multicolumn{1}{c|}{64.9\%} & \multicolumn{1}{c|}{57.4\%} & 55.3\% & \multicolumn{1}{c|}{86.1\%} & \multicolumn{1}{c|}{55.0\%} & 47.6\% & \multicolumn{1}{c|}{86.1\%} & \multicolumn{1}{c|}{60.3\%} & \multicolumn{1}{c|}{50.2\%} & 47.6\% \\ \hline
Llama-3.1-70B-Instruct & 57.0\% & \multicolumn{1}{c|}{56.6\%} & 57.3\% & \multicolumn{1}{c|}{64.0\%} & \multicolumn{1}{c|}{51.6\%} & 51.3\% & \multicolumn{1}{c|}{82.0\%} & \multicolumn{1}{c|}{52.1\%} & 45.1\% & \multicolumn{1}{c|}{82.0\%} & \multicolumn{1}{c|}{57.7\%} & \multicolumn{1}{c|}{48.8\%} & 45.1\% \\ \hline
Qwen2-72B-Instruct & 55.7\% & \multicolumn{1}{c|}{51.5\%} & 58.8\% & \multicolumn{1}{c|}{61.4\%} & \multicolumn{1}{c|}{52.6\%} & 49.2\% & \multicolumn{1}{c|}{80.4\%} & \multicolumn{1}{c|}{50.1\%} & 46.9\% & \multicolumn{1}{c|}{80.4\%} & \multicolumn{1}{c|}{56.3\%} & \multicolumn{1}{c|}{45.5\%} & 46.9\% \\ \hline
Llama-3-70B-Instruct & 53.7\% & \multicolumn{1}{c|}{52.3\%} & 54.8\% & \multicolumn{1}{c|}{60.1\%} & \multicolumn{1}{c|}{48.6\%} & 48.5\% & \multicolumn{1}{c|}{78.9\%} & \multicolumn{1}{c|}{47.8\%} & 46.1\% & \multicolumn{1}{c|}{78.9\%} & \multicolumn{1}{c|}{51.5\%} & \multicolumn{1}{c|}{44.4\%} & 46.1\% \\ \hline
Qwen1.5-72B-Chat & 47.5\% & \multicolumn{1}{c|}{46.1\%} & 48.5\% & \multicolumn{1}{c|}{51.6\%} & \multicolumn{1}{c|}{45.1\%} & 43.2\% & \multicolumn{1}{c|}{75.2\%} & \multicolumn{1}{c|}{41.2\%} & 37.5\% & \multicolumn{1}{c|}{75.2\%} & \multicolumn{1}{c|}{42.7\%} & \multicolumn{1}{c|}{40.1\%} & 37.5\% \\ \hline
Llama-3.1-8B-Instruct & 44.1\% & \multicolumn{1}{c|}{44.0\%} & 44.2\% & \multicolumn{1}{c|}{49.0\%} & \multicolumn{1}{c|}{43.3\%} & 36.4\% & \multicolumn{1}{c|}{70.0\%} & \multicolumn{1}{c|}{38.3\%} & 34.9\% & \multicolumn{1}{c|}{70.0\%} & \multicolumn{1}{c|}{41.0\%} & \multicolumn{1}{c|}{35.6\%} & 34.9\% \\ \hline
Qwen2-7B-Instruct & 41.6\% & \multicolumn{1}{c|}{40.5\%} & 42.4\% & \multicolumn{1}{c|}{45.6\%} & \multicolumn{1}{c|}{40.7\%} & 35.5\% & \multicolumn{1}{c|}{70.9\%} & \multicolumn{1}{c|}{34.1\%} & 35.1\% & \multicolumn{1}{c|}{70.9\%} & \multicolumn{1}{c|}{35.8\%} & \multicolumn{1}{c|}{32.4\%} & 35.1\% \\ \hline
Qwen1.5-14B-Chat & 40.2\% & \multicolumn{1}{c|}{38.6\%} & 41.3\% & \multicolumn{1}{c|}{44.1\%} & \multicolumn{1}{c|}{38.4\%} & 35.3\% & \multicolumn{1}{c|}{73.8\%} & \multicolumn{1}{c|}{32.4\%} & 28.9\% & \multicolumn{1}{c|}{73.8\%} & \multicolumn{1}{c|}{32.2\%} & \multicolumn{1}{c|}{30.9\%} & 28.9\% \\ \hline
Llama-3-8B-Instruct & 39.9\% & \multicolumn{1}{c|}{39.8\%} & 40.0\% & \multicolumn{1}{c|}{45.0\%} & \multicolumn{1}{c|}{37.0\%} & 34.4\% & \multicolumn{1}{c|}{68.4\%} & \multicolumn{1}{c|}{34.0\%} & 27.5\% & \multicolumn{1}{c|}{68.4\%} & \multicolumn{1}{c|}{36.1\%} & \multicolumn{1}{c|}{31.6\%} & 27.5\% \\ \hline
Mistral-7B-Instruct-v0.3 & 35.2\% & \multicolumn{1}{c|}{35.2\%} & 35.1\% & \multicolumn{1}{c|}{40.5\%} & \multicolumn{1}{c|}{31.6\%} & 30.1\% & \multicolumn{1}{c|}{71.7\%} & \multicolumn{1}{c|}{25.8\%} & 27.2\% & \multicolumn{1}{c|}{71.7\%} & \multicolumn{1}{c|}{25.0\%} & \multicolumn{1}{c|}{23.4\%} & 27.2\% \\ \hline
Qwen1.5-7B-Chat & 34.4\% & \multicolumn{1}{c|}{33.6\%} & 35.0\% & \multicolumn{1}{c|}{40.1\%} & \multicolumn{1}{c|}{29.9\%} & 29.7\% & \multicolumn{1}{c|}{69.3\%} & \multicolumn{1}{c|}{25.1\%} & 28.3\% & \multicolumn{1}{c|}{69.3\%} & \multicolumn{1}{c|}{24.8\%} & \multicolumn{1}{c|}{22.9\%} & 28.3\% \\ \hline
Deepseek-Coder-7B-Instruct-v1.5 & 34.1\% & \multicolumn{1}{c|}{33.9\%} & 34.2\% & \multicolumn{1}{c|}{38.4\%} & \multicolumn{1}{c|}{33.0\%} & 27.6\% & \multicolumn{1}{c|}{68.4\%} & \multicolumn{1}{c|}{25.1\%} & 26.5\% & \multicolumn{1}{c|}{68.4\%} & \multicolumn{1}{c|}{22.7\%} & \multicolumn{1}{c|}{24.2\%} & 26.5\% \\ \hline
Gemma-7B-Instruct & 23.3\% & \multicolumn{1}{c|}{20.6\%} & 25.3\% & \multicolumn{1}{c|}{28.7\%} & \multicolumn{1}{c|}{18.2\%} & 19.9\% & \multicolumn{1}{c|}{48.2\%} & \multicolumn{1}{c|}{15.7\%} & 22.9\% & \multicolumn{1}{c|}{48.2\%} & \multicolumn{1}{c|}{13.5\%} & \multicolumn{1}{c|}{14.9\%} & 22.9\% \\ \hline
Tablellama & 21.1\% & \multicolumn{1}{c|}{23.9\%} & 19.1\% & \multicolumn{1}{c|}{25.4\%} & \multicolumn{1}{c|}{20.0\%} & 14.9\% & \multicolumn{1}{c|}{45.4\%} & \multicolumn{1}{c|}{16.3\%} & 9.9\% & \multicolumn{1}{c|}{45.4\%} & \multicolumn{1}{c|}{15.5\%} & \multicolumn{1}{c|}{14.4\%} & 9.9\% \\ \hline
\end{tabular}
\caption{Performance of LLMs on MiMoTable. GPT-4o-CI refers to the GPT-4o model with a code interpreter plugin.}
\label{result_of_our}
\end{table*}

\begin{figure*}[h]
\centering
\includegraphics[width=1\linewidth]{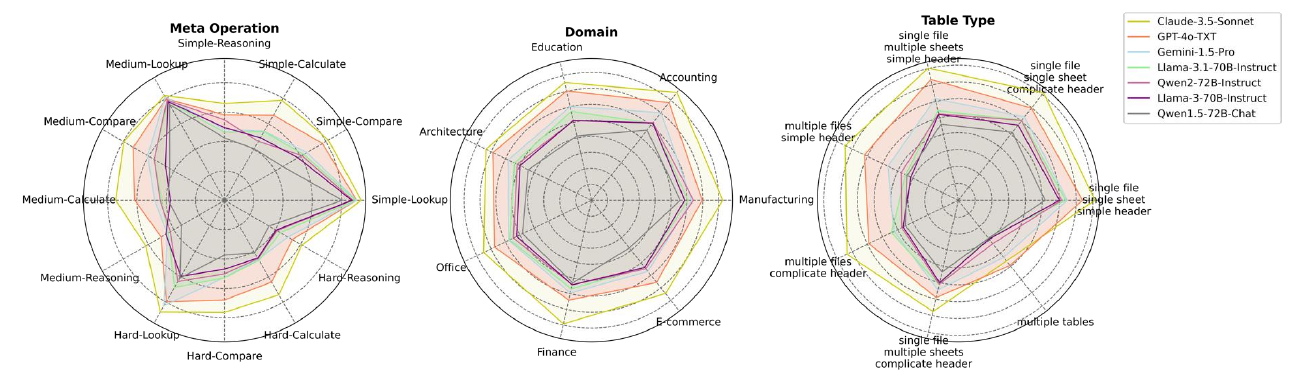}
\caption{The performance of LLMs on MiMoTable respecting to different Meta Operations, Domains and Table Types.}
\label{radar}
\end{figure*}

\section{Experiments and Results}
Our experiments have two main goals: (1) to evaluate representative LLMs' performance on our dataset; and (2) to prove the effectiveness of proposed meta operations. This section presents the relevant experiments and findings.

\subsection{Experimental Setup}
\noindent\textbf{Models.} We conducted experiments on 16 selected LLMs, comprising open-source LLMs, closed-source LLMs, and tabular LLMs. The open-source LLMs we evaluated include Llama3.1\footnote{https://ai.meta.com/blog/meta-llama-3-1/}, Llama3 \cite{llama3}, Qwen2 \cite{qwen2}, Qwen1.5 \cite{qwen}, Mistral \cite{mistral}, DeepseekCoder \cite{deepseekcoder}, and Gemma \cite{gemma}. The closed-source LLMs are GPT-4o \cite{gpt4}, Claude-3.5-Sonnet\footnote{https://www.anthropic.com/news/claude-3-5-sonnet}, and Gemini-1.5-Pro \cite{gemini1.5}. We also evaluated Tablellama \cite{TableLlama}, a tabular model fine-tuned specifically for various table tasks. However, most tabular models, such as Binder \cite{Binder}, require inputs to be tables with known headers, which is not suitable for our benchmark.

\noindent\textbf{Datasets.} To demonstrate the generality and effectiveness of our meta operation, we conducted experiments on the newly proposed benchmark as well as two existing open-source benchmarks: WikiTableQuestion and WikiSQL. These widely-used TableQA benchmarks feature tables sourced from Wikipedia with simple headers.

\noindent\textbf{Metrics.} We used accuracy as the evaluation metric. Except for Tablellama, the predicted answers of other models in our experiments are all free-formed. We prompted GPT-4o to judge the correctness of the predicted answer based on the question and human-verified reference answer. Because a small portion of questions in MiMoTable are open-ended, we also instructed GPT-4o to give a score between 0-1 when it judges the question is open-ended. 

\noindent\textbf{Implementation Details.} For all LLMs except Tablellama, we input table contents in markdown format. For GPT-4o, we also tested another popular approach in table reasoning, which is denoted as GPT-4o-CI in Table \ref{result_of_our}.  This method uploads spreadsheets and generates Python code, executed via a code interpreter plugin, with results fed back for analysis. The format of MiMoTable and WikiTableQuestion is spreadsheet files, which can be directly as part of inputs to GPT-4o-CI. For WikiSQL, we first saved the non-file table content as Excel files and fed them to GPT-4o-CI. For Tablellama, we followed the prompt format specified in the original paper. For the existing benchmarks, we used GPT-4o to divide the questions according to the meta operations and then calculated the difficulty scores of the datasets based on Equation \ref{score of dataset}. We use the official default parameters for all models. More details can be found in the supplementary material.

\begin{figure*}[h]
\centering
\includegraphics[width=0.97\linewidth]{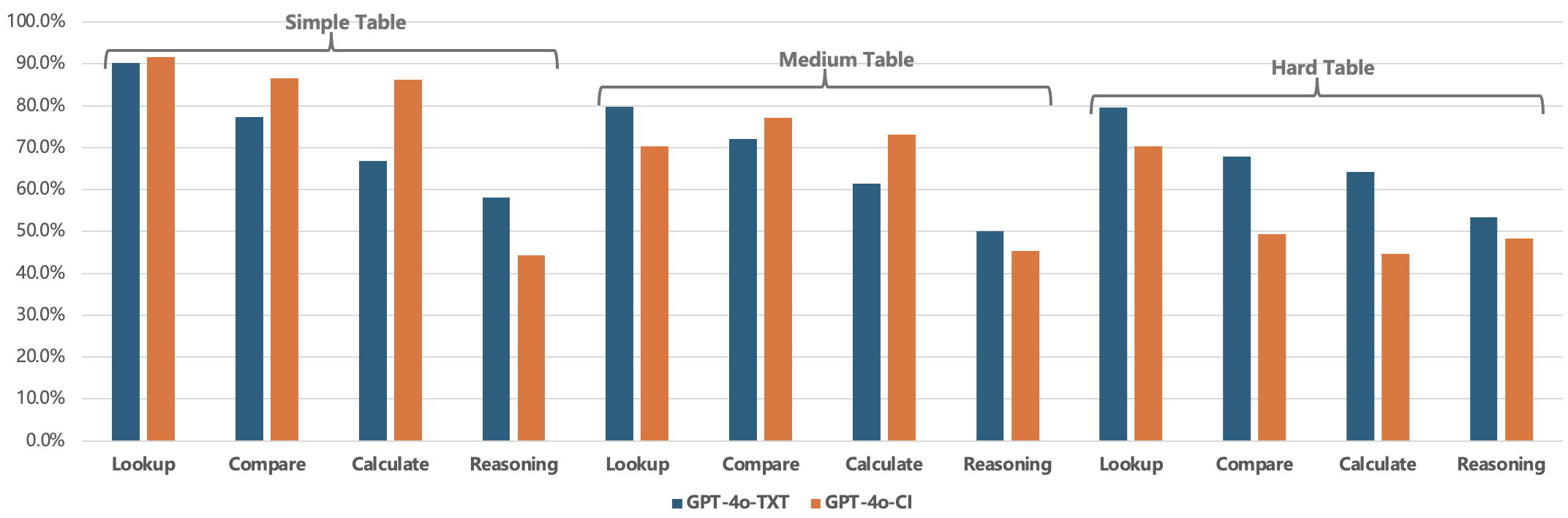}
\caption{Performance on different data types. }
\label{exp_two_approach}
\end{figure*}

\subsection{Results and Analysis}
\noindent\textbf{Overall Performance.} As shown in Table \ref{result_of_our}, we evaluate our proposed benchmark dataset using different LLMs. Because most experimental LLMs can not generate edited files and charts, we only infer the questions without meta operations Edit and Visualize for a fair comparison with GPT-4o-CI. As we can see, the best-performing model, Claude-3.5-Sonnet, achieved an overall performance of only 77.4\% on our benchmark, highlighting that MiMoTable poses significant challenges for current LLMs. This underscores the need for further exploration to improve model performance on more realistic table data.

\noindent\textbf{Analysis of different approaches.} There are mainly two approaches to solving table reasoning problems of spreadsheets. One approach is to represent the spreadsheet content in text form and input it into the model to directly generate answers. The other is directly using the spreadsheet as input to write code, run in a sandbox, and conclude to solve the problem in a ReAct \cite{ReAct} way. As shown in Figure \ref{exp_two_approach}, we compare the performance of those two methods based on the GPT-4o model, where GPT-4o-TXT is the first text form approach and GPT-4o-CI is the second code-based approach. The GPT-4o-CI performs better than GPT-4o-TXT in Calculate and Compare when the table difficulty is simple and medium, while GPT-4o-TXT performs better in hard tables and in meta operations of Lookup and Reasoning. This reveals that the code-based approach has advantages in calculating only when the tables are not so hard, as hard tables can cause the model to be unable to write the correct code to locate the required data. The text-based approach is good at Lookup and Reasoning because the model can see the entire content of the table as long as the context window size is enough. The first radar chart in Figure \ref{radar} shows the results of more LLMs respecting to the different combinations of table difficulty and meta operations.

\begin{figure*}[h]
\centering
\includegraphics[width=1\linewidth]{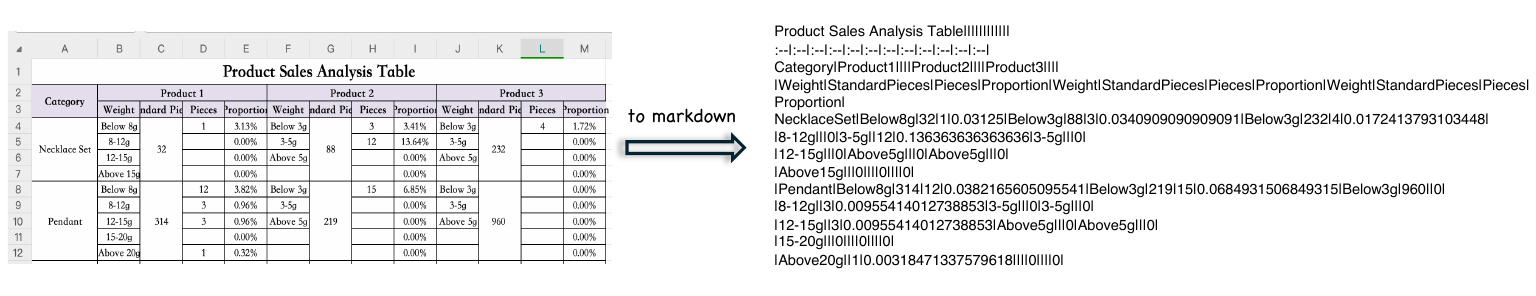}
\caption{A spreadsheet to text with the markdown format.}
\label{markdown_trans}
\end{figure*}

\noindent\textbf{Capability of LLMs for table reasoning.} According to the results of table difficulty in Table \ref{result_of_our} and table types in Figure \ref{radar}, most LLMs have struggled in medium and hard tables. We attribute the reasons to two factors, hierarchical header and multiple similar tables. As illustrated in Figure \ref{markdown_trans}, although the hierarchical relations between table cells appear very clear in the original spreadsheets, they become much less intuitive when converted into text, which poses challenges for the LLMs to understand.
Additionally, the tables in the spreadsheets with multiple sheets are usually very similar. LLMs need to comprehensively consider multiple similar tables to answer questions. In conclusion, the ability to understand complex table structures and multiple similar tables in table reasoning needs to be improved for current LLMs.
We also observe differences in the performance of different models across languages. For example, Claude-3.5-Sonnet performs better in English than in Chinese, while Qwen2-72B-Instruct is the opposite. We believe this is due to the varying proportions of different languages used during the pretraining and SFT  stages for each model.

\begin{figure}[h]
\centering
\includegraphics[width=1\columnwidth]{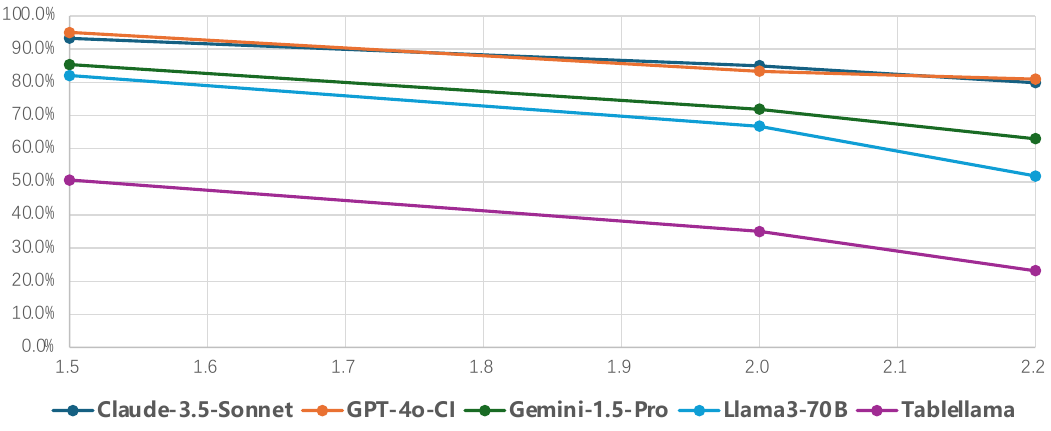}
\caption{The relations between performance and the difficulty of benchmarks. The x-axis is the difficulty of benchmarks graded by meta operations. The y-axis is the accuracy of tested LLMs.}
\label{relations_of_preformance_and_difficulty}
\end{figure}

\noindent\textbf{Effectiveness of Meta Operations.} Although WikiTableQuestion and WikiSQL are both datasets for TableQA tasks with tables that have simple headers, we found that the performance of the same LLMs on these two datasets varies significantly. For example, Llama3-70B achieves 82.0\% accuracy on WikiSQL but only 66.7\% accuracy on WikiTableQuestion. No objective metric exists to explain this discrepancy. By scoring the datasets according to our meta operations, we found that WikiTableQuestion is significantly more difficult than WikiSQL: the difficulty of WikiSQL is 1.5, while the difficulty of WikiTableQuestion is 2.0. The questions involving simple tables in MiMoTable, denoted as MiMoTable-Simple, have a difficulty of 2.2. We evaluated the performance of different LLMs on these three datasets—WikiSQL, WikiTableQuestion, and MiMoTable-Simple—and the results are shown in Figure \ref{relations_of_preformance_and_difficulty}. The x-axis represents the difficulty of the benchmarks graded by meta operations, and the y-axis shows the accuracy of the LLMs on these benchmarks. We found that as the difficulty score of the dataset increases, the model performance declines. This indicates that our proposed meta operations and difficulty scores are both generalizable and effective across different benchmarks.

\section{Related Work}
The main tasks for table reasoning include four categories: TableQA, Table2Text, Table Manipulation, and Advanced Data Analysis \cite{Survey}. Researchers have proposed various table benchmarks for these tasks. TableQA is the most popular task, including benchmarks like
WikiTableQuestions \cite{WikiTableQuestion}, WikiSQL \cite{WikiSQL}, FeTaQA \cite{FetaQA}, HybridQA \cite{HybridQA}, TATQA \cite{TAT-QA}, NQ-TABLES \cite{NQ-TABLES}, HybriDialogue \cite{HybriDialogue}, BIRD \cite{BIRD}, Spider \cite{Spider}. The primary benchmark for Table2Text is ToTTo \cite{ToTTo}. A high-quality Table Manipulation benchmark called WikiTableEdit is introduced in \cite{WikiTableEdit}. SPREADSHEETBENCH \cite{spreadsheetbench} is a challenging spreadsheet manipulation benchmark. 
For the Advanced Data Analysis task, the benchmarks DAEval \cite{InfiAgent-DABench} and DS-1000 \cite{DS-1000} are proposed. Text2Analysis \cite{Text2Analysis} is a recently introduced benchmark that includes both TableQA and Advanced Data Analysis tasks. Most existing table benchmarks feature simple table headers, but HiTab \cite{HiTab} is a TableQA and Table2Text dataset based on hierarchical headers. AIT-QA \cite{AIT-QA} is a dataset for TableQA with hierarchical headers specific to the airline industry.

Unlike the existing benchmarks, we propose a new benchmark, MiMoTable, the first benchmark with multi-scale spreadsheets that simultaneously covers four tasks: TableQA, Table2Text, Table Manipulation, and Advanced Data Analysis.

\section{Conclusion}
We propose a multi-scale spreadsheet benchmark with four tasks: TableQA, Table2Text, Table Manipulation, and Advanced Data Analysis, named MiMoTable. Experiments have shown that existing LLMs perform poorly on this benchmark, indicating that there is still significant room to improve in more realistic scenarios. For table reasoning, we also propose a new criterion for categorizing problems based on meta operations. Compared to task-based categorization, this criterion allows for a deeper and more accurate analysis of problems in table datasets. Our experiments demonstrate that the meta operations are general and effective.

\section{Limitations}
When validating the effectiveness of meta operations, we do not perform Supervised Fine-Tuning (SFT) on the models. Future work could examine the role and effect of each type of operation through SFT. Meanwhile, we used the same prompt for evaluating all models except Tablellama, without optimizing or adapting prompts for different models. Regarding hyperparameters for different models, we used the officially recommended default parameters and do not adjust different hyperparameters for different models.
Additionally, inspired by work in other fields \cite{icl, cs}, developing long-context table reasoning benchmarks and studying in-context learning for table reasoning are valuable directions for further exploration.

\section*{Acknowledgments}
We thank the three anonymous reviewers for carefully reading our paper and their insightful comments and suggestions.
\bibliography{custom}
\clearpage
\appendix

\section{Appendix}
\label{sec:appendix}

\subsection{Data Statistic of The Language}
Table \ref{stat_of_lan} shows the data statistics of the proposed benchmark under different languages, including table number, question number, and the average of question difficulty.

\begin{table}[h]
\centering
\scriptsize
\renewcommand\tabcolsep{5pt}
\renewcommand\arraystretch{1.2}
\begin{tabular}{c|c|c|c} \hline
     & \textbf{Table Number} & \textbf{Question Number} & \textbf{Question Difficulty} \\ \hline
    \textbf{Overall} & 428 & 1719 & 2.2\\ \hline
    \textbf{English} & 182 & 671 & 2.2\\ \hline
    \textbf{Chinese} & 246 & 1048 & 2.2\\ \hline
\end{tabular}
\caption{Data Statistics of Different Languages}
\label{stat_of_lan}
\end{table}

\subsection{Used Prompts}
Table~\ref{prompt_for_op}, Table~\ref{prompt_for_infer}, and Table~\ref{prompt_for_eval} show the designed prompts for meta operations classification, model inference, and performance evaluation in this paper.

\begin{table}[h]
\centering
\small
\renewcommand\arraystretch{1.3}
\begin{tabular}{p{7cm}}
\hline
\textit{You are a spreadsheet question classification expert. Given a user's question about an Excel spreadsheet, classify the question according to the requirements and output it in the specified format.} \\
\\
\textless{}Requirements\textgreater{} \\
The following operation classification already exists, presented in the format of operation name: operation description. If the user's question can be classified as some of the operations, output the operation names. One question can be classified into multiple operations. \\
Lookup: Locate the position of the specific target. \\
Edit: Modify, delete, or add to a table. \\
Calculate: The numerical computation, sum, avg, max, etc. \\
Compare: Compare two or more targets in a table. \\
Visualize: Show in chart.
Reasoning: Inferring information from the table content that is not explicitly included. \\
\textless{}/Requirements\textgreater{} \\
\textless{}Output Format\textgreater{} \\
operation name 1, operation name 2, ... \\
\textless{}/Output Format\textgreater{}\\ 
\\
\textless{}Question\textgreater{} \\
what country hosted the most tournaments? \\
\textless{}/Question\textgreater{} \\
Lookup, Calculate, Compare \\
\\
\textless{}Question\textgreater{} \\
QUESTION TO BE CLASSIFIED \\
\textless{}/Question\textgreater{} \\ \hline

\end{tabular}
\caption{Prompt for Meta Operation Classification}
\label{prompt_for_op}
\end{table}

\begin{table}[t!]
\centering
\small
\renewcommand\arraystretch{1.3}
\begin{tabular}{p{7cm}}
\hline
\textit{Below is the table content in markdown, please answer the question according to the table content.} \\ \\ \textless{}Table\textgreater{} \\
\textless{}Table Name\textgreater{} \\
SPREADSHEET FILE NAME \\ 
\textless{}/Table Name\textgreater{} \\ 
\textless{}Table Content\textgreater{} \\
\textless{}Sheet\textgreater{} \\ 
\textless{}Sheet Name\textgreater{}\\
SHEET NAME\\
\textless{}/Sheet Name\textgreater{}\\
\textless{}Sheet Content\textgreater{} \\
SHEET CONTENT IN MARKDOWN \\
\textless{}/Sheet Content\textgreater{} \\
\textless{}/Sheet\textgreater{} \\
\textless{}/Table Content\textgreater{} \\
\textless{}/Table\textgreater{}  \\ \\
\textless{}Question\textgreater{} \\
THE QUESTION TO BE ASKED\\
\textless{}/Question\textgreater{} \\
\hline
\end{tabular}
\caption{Prompt for Model Inference}
\label{prompt_for_infer}
\end{table}

\begin{table}[t!]
\centering
\small
\renewcommand\arraystretch{1.3}
\begin{tabular}{p{7cm}}
\hline

\textit{For the following questions, given the correct answer, determine whether the candidate's answer is correct. If it is correct, output "Correct"; if it is incorrect, output "Incorrect";
if it is uncertain whether it is correct, output "Uncertain".
As long as the candidate's answer contains the key information that can correctly answer the question, it is considered correct. If the question is open-ended, give a score between 0-1 according to the correct answer. Do not output any other content.} \\ \\

\textless{}Question\textgreater{}

THE QUESTION

\textless{}/Question\textgreater{} \\

\textless{}Correct Answer\textgreater{}

THE CORRECT ANSWER

\textless{}/Correct Answer\textgreater{} \\

\textless{}Candidate answer\textgreater{}

THE CANDIDATE ANSWER

\textless{}/Candidate answer\textgreater{}  \\
\hline
\end{tabular}
\caption{Prompt for Performance Evaluation}
\label{prompt_for_eval}
\end{table}

\end{document}